# Recognition of cDNA microarray image Using Feedforward artificial neural network

R. M. Farouk[1] , S. Badr[2] and M. Sayed Elahl[2]

[1]Department of Mathematics, Faculty of Science, Zagazig University,

P. O. Box 44519, Zagazig, Egypt. E-mail: rmfarouk1@yahoo.com

[2] Department of Mathematics, Faculty of Science, Banha University, Egypt. E-mail:

ma0777@yahoo.com

## Abstract

The complementary DNA (cDNA) sequence is considered to be the magic biometric technique for personal identification. In this paper, we present a new method for cDNA recognition based on the artificial neural network (ANN). Microarray imaging is used for the concurrent identification of thousands of genes. We have segmented the location of the spots in a cDNA microarray. Thus, a precise localization and segmenting of a spot are essential to obtain a more accurate intensity measurement, leading to a more precise expression measurement of a gene. The segmented cDNA microarray image is resized and it is used as an input for the proposed artificial neural network. For matching and recognition, we have trained the artificial neural network. Recognition results are given for the galleries of cDNA sequences . The numerical results show that, the proposed matching technique is an effective in the cDNA sequences process. We also compare our results with previous results and find out that, the proposed technique is an effective matching performance.

## 1. Introduction

Biometrics refers to a distinctive, measurable characteristics automated system that can identify an individual by measuring their characteristics or traits or patterns, and comparing it to those on record. The two categories of biometric identifiers include physiological and behavioral characteristics. A biometric would identify ones by iris, fingerprint, voice, cDNA, hand print and signature or behavior [1].

Deoxyribonucleic acid (DNA) biometrics could be the most exact form of identifying any given individual. Every human being has its own individual map for every cell made, and this map can be found in everybody cell. Because DNA is the structure that defines who we are physically and intellectually, unless an individual is an identical twin, it is not likely that any other person will have the same exact set of genes [2].

Molecular biologists and Bioinformatics are using microarray technology for identifying a gene in a biological sequence and predicting the function of the identified gene within a larger system. Microarray technology is based on creating DNA microarrays that are typically composed of thousands of DNA sequences, called probes, fixed to a glass or silicon substrate [3,4].

Rapid progress has been made in improving the recognition rates. This development has greatly improved the scope of using cDNA for identification purposes. The current works focus on enhancing the feature extraction and matching and improving the accuracy of the results. Today, with the crime rates increasing everyday there is an urgent need for a system which is safe and fast. The general cDNA recognition system based on feature extraction and classification of the image using the features extracted.

cDNA matching has become a popular use in criminal trials, especially in proving rape cases. The main problems surrounding cDNA biometrics is that it is not a quick process to identify someone by their cDNA. The process is also a very costly one [5]. Applications of DNA microarrays are becoming a common tool in many areas of microbial research, including microbial physiology, pathogenesis, epidemiology, ecology, phylogeny, pathway engineering and fermentation optimization [6]. Different DNA based molecular techniques viz. PCR, AFLP, RAPD, RFLP, PCR,SSCP, real time PCR, DNA microarray have been used for identification and to assess the genetic diversity among the parasite population [7].

Microarrays can be constructed with dozens to millions of probes on their surface to allow high throughput analyses of many biologic processes to be performed simultaneously on the same sample. It enables systematic surveys of variations in DNA sequence and gene expression. Microarrays are now widely used for gene expression analysis, deoxyribonucleic acid resequencing, single-nucleotide polymorphism genotyping, and comparative genomic hybridization [8].

Automatic grid alignment together with the FPGA based hardware architectures are an efficient solution for fast and automated microarray image processing, overcoming the disadvantages of existing software platforms in case of a large number of microarray analyses are needed [9].

### 1. 2 Outline

The rest of this paper is organized as follows: in section 2 we have discussed cDNA microarray images preprocessing; enhancing, gridding, segmentation, in section 3, we have used the segmented microarray images as an input of ANN. This method involves the training of the entire pattern at once against the other earlier works of training and in section 4, we have discussed our results. The overall algorithm is shown in Fig 1.

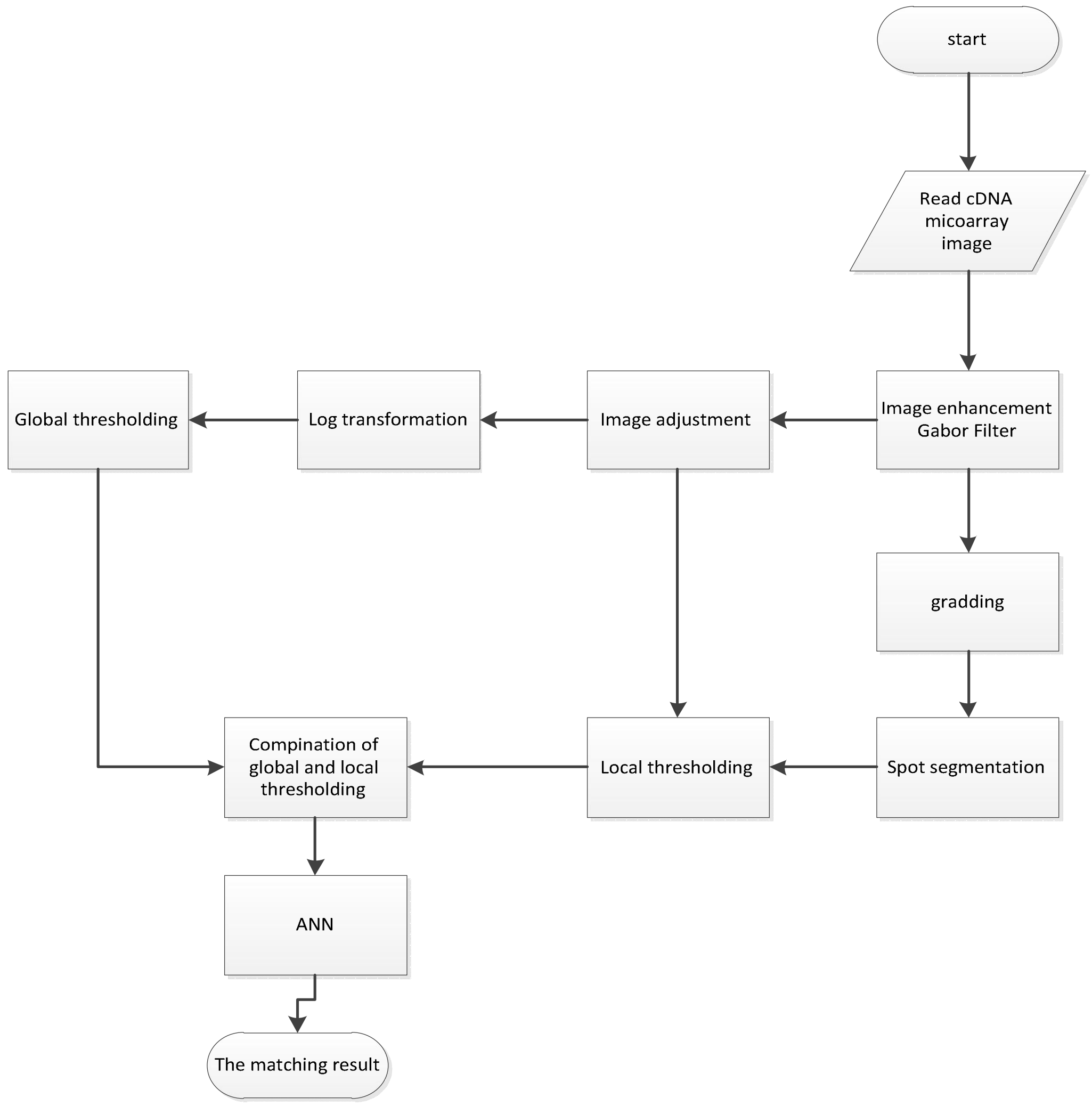

Fig. 1 Flowchart of the proposed technique

## 2. cDNA image enhancement

Microarray imaging is considered an important tool for large scale analysis of gene expression. Microarrays are arrays of glass microscope slides, in which thousands of discrete DNA sequences are printed by a robotic array, thus, forming circular spots of known diameter. It consists of thousands of spots representing chains of cDNA sequences, arranged in a two-dimensional array [10].

The accuracy of the gene expression depends on the experiment itself and further image processing. It's well known that the noises introduced during the experiment will greatly affect the accuracy of the gene expression, so there are many studies and approach to enhance and reduce the noise on cDNA microarray images. CDNA microarray image analysis is a new and exciting computing paradigm with biological background. It involves the concepts ranging from molecular biology, statistics to computation while include theories and practices of computer science. Study in DNA microarray includes the gene expression, genetic data management, repository system and the algorithms of image analysis [11].

Microarray image enhancement by denoising using stationary wavelet transform provides a better performance in denoising than traditional wavelet transform method [12].

Removing noise impairments while preserving structural information in cDNA microarray images using fuzzy vector filtering framework, Noise removal is performed by tuning a membership function which utilizes distance criteria applied to cDNA vectorial inputs at each image location [13].

The task in microarray image enhancement involves removing the image background, noise from scanning and illustrating the guide spots of the cDNA [14]. It typically involves the following steps:

(i) Filter cDNA Image and get the guide spots (using Gabor filter).

(ii) Identify the location of all spots on the microarray image.

(iii) Generates the grid within each block which subdivides the block into $n \times m$ sub regions, each containing at most one spot.

(iv) Segment the spot, if any, in each sub region.

## 2.1 Gabor filtering

The Gabor filter method has been widely used both in multi-resolution image processing and feature extraction for object identification [15]. Given an image with size M x N, a 2D Gabor function is a Gaussian modulated by a sinusoid. It is a non-orthogonal wavelet and it can be specified by the frequency of the sinusoid and the standard deviations of Gaussian x and y [16]:

$$G k = \frac{k^2}{\sigma^2}\left\{ \exp^{-\left(\frac{x^2+y^2}{2\sigma^2}\right)\left\|k^2\right\|}\left( \exp^{i\vec{k}\vec{x}} - \exp^{\frac{\sigma^2}{2}} \right) \right. \tag{1}$$

$$\vec{x} = (x, y) \quad pixel\ coordinates$$
$$\vec{k} = k\cos\varphi + k\sin\varphi$$
$$\vec{x}.\vec{k} = k\cos\varphi\, x + k\sin\varphi\, y$$

Euler form Substitute

$$\exp(ikx) = \underbrace{\cos\left(x * k\cos\varphi + y * k\sin\varphi\right)}_{\text{Real part}} + i\,\underbrace{\boxed{\sin\left(x * k\cos\varphi + y * k\sin\varphi\right)}}_{\text{Imaginary part}}$$

$$Gkreal\ part = \frac{k^2}{\sigma^2}\left\{ \exp^{-\left(\frac{x^2+y^2}{2\sigma^2}\right)\left\|k^2\right\|} \underline{\cos\left(x * k\cos\varphi + y * k\sin\varphi\right)} \right.$$

$$Gk\ img\ part = \frac{k^2}{\sigma^2}\left\{ \exp^{-\left(\frac{x^2+y^2}{2\sigma^2}\right)\left\|k^2\right\|} \boxed{\sin\left(x * k\cos\varphi + y * k\sin\varphi\right)} \right.$$

The Gabor filter used to enhance the image and Fig. 2.

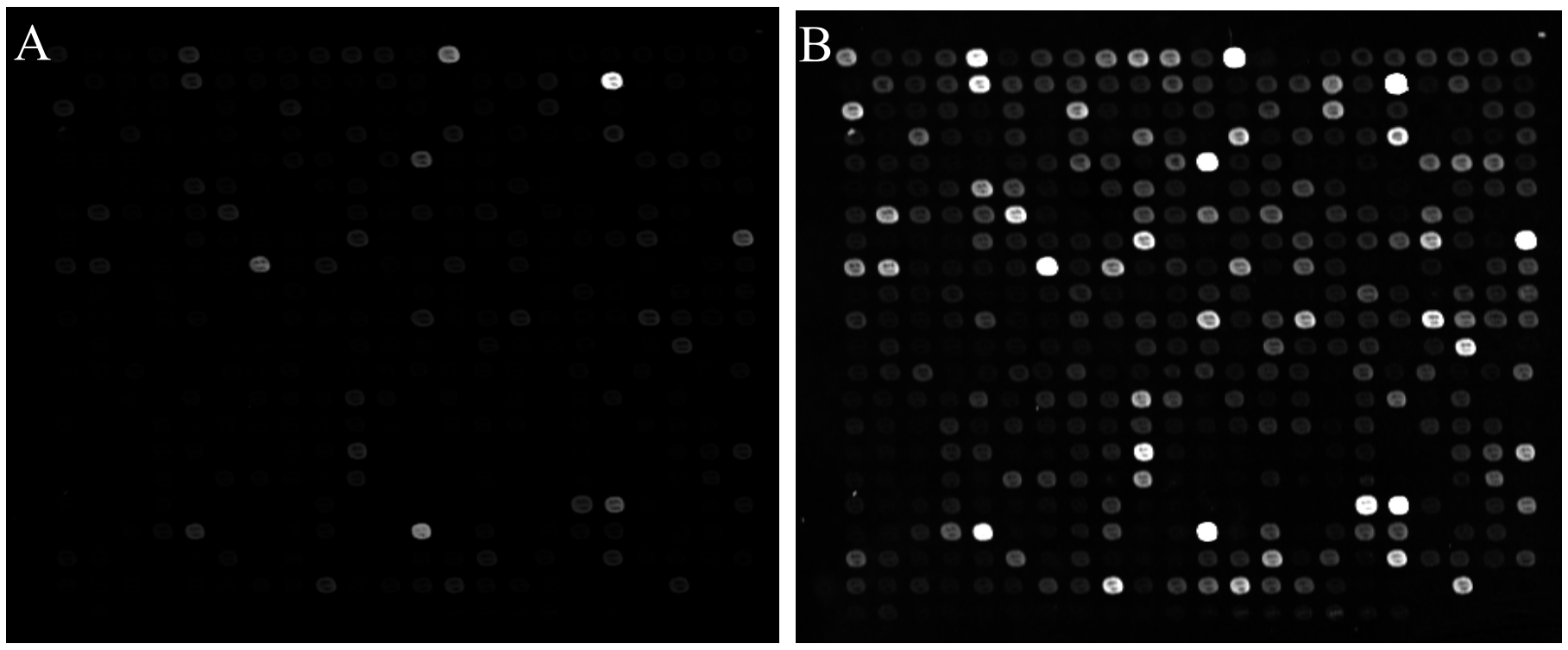

Fig. 2 The effect of Gabor filter on (A) original image, (B) Gabor filtered image

## 2.2 Gridding

Images from microarray experiments are highly structured since they are composed of high intensity spots located on a regular grid. The shape of the spots is roughly circular, although variations are possible. The ideal microarray image has the following properties [17,18]

(i) All the sub grids are about the same size.

(ii) The spacing between sub grids is regular.

(iii) The location of the spots is centered on the intersections of the lines of the sub grid.

(iv) The size and shape of the spots are perfectly circular and it is the same for all the spots.

(v) The location of the grids is fixed in images for a given type of slides.

(vi) No dust or contamination is on the slide.

(vii) There is minimal and uniform background intensity across the image.

It goes without saying that almost all real microarray images violate at least one of these conditions. In fact, we frequently observed variations on the spot position, irregularities on the spot shape and size, contamination, and global problems that affect multiple spots. In general, the shape and the size of the spots may fluctuate significantly across the array. The microarray image gridding is the separation process of the spots into distinct cells, to generate the correct grid for the microarray image, consider the location of good quality spots (which act as guide spots). A good-quality spot should be circular in shape, of appropriate size, and with intensity consistently higher than the background. Moreover, its position should agree with the overall spot geometry as dictated by the printing process. After the guide spots are found, the correct grid can be generated based on their geometry [19,20], Fig. 4. To find the grid lines in the microarray images is based on a series of steps:

(i) Create a horizontal profile to get the mean intensity for each column of the image.

(ii) Autocorrelation to enhance the self-similarity of the profile. The smooth result promotes peak finding and estimation of spot spacing Fig. 3.

(iii) Locate centers of the spots.

(iv) Determine divisions between spot, The midpoints between adjacent peaks provides grid point locations.

(v)Transpose the image and repeat the previous steps to get the vertical grid.

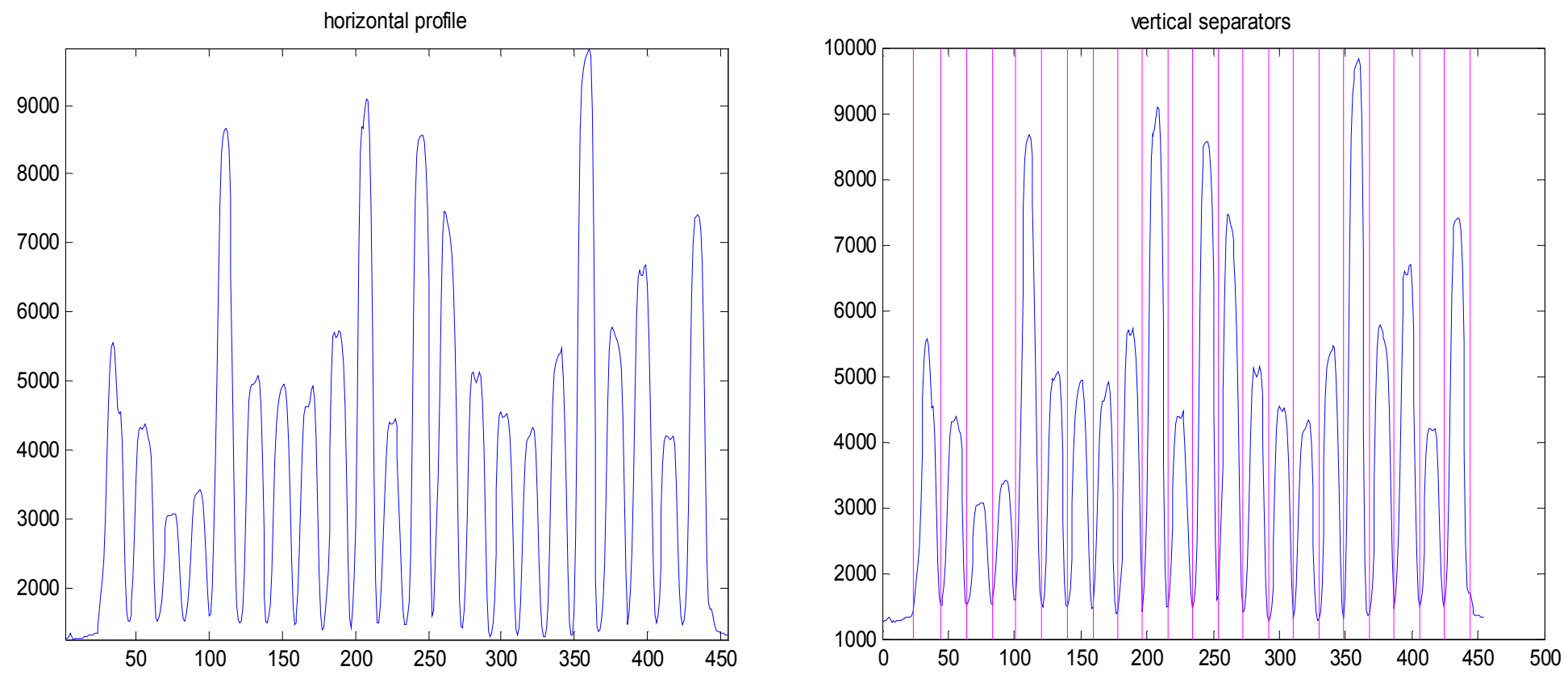

Fig. 3 Steps of gridding

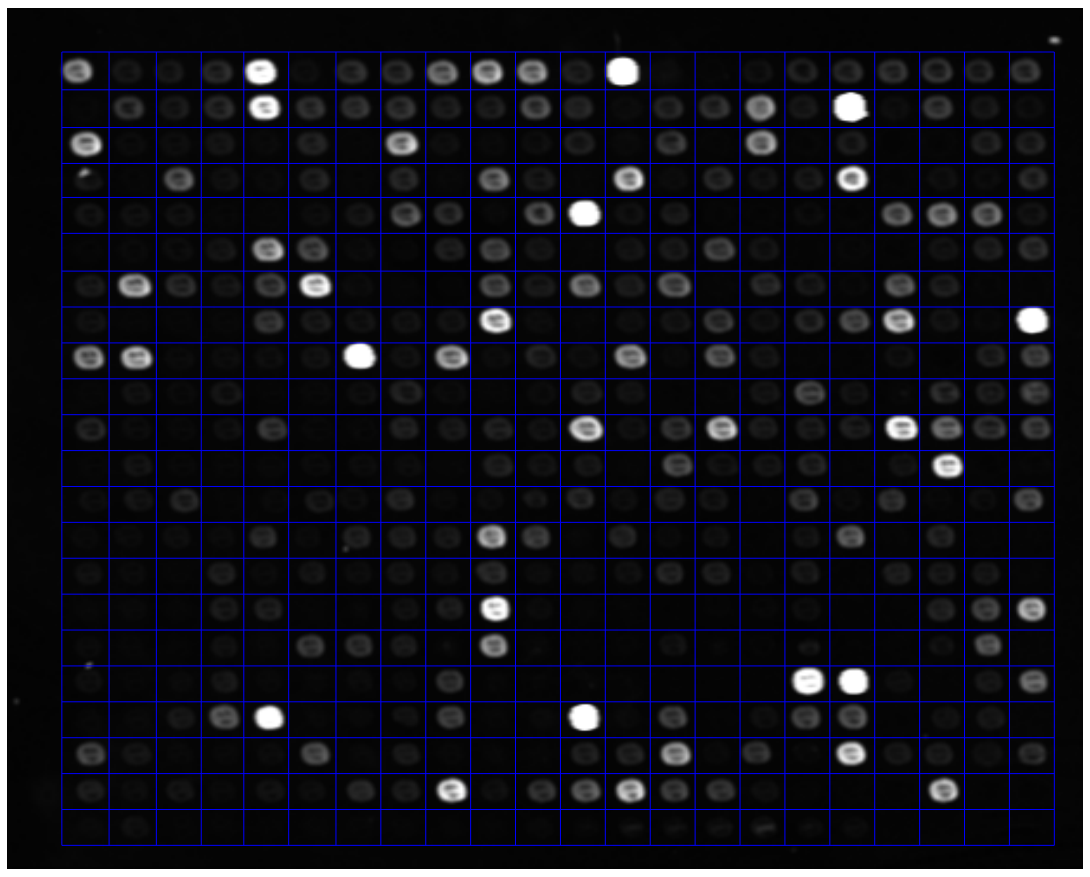

Fig. 4 Result of automatic gridding

### 2.3 Segmentation

Image segmentation is the process of distinguishing objects from the background and partitioning the image into several regions, each having its own properties. It is usually the first step in vision systems, and is the basis for further processing such as description or recognition. In microarray image processing, segmentation refers to the classification of pixels as either the signal or the surrounding area. Most image segmentation approaches can be placed in one of five categories [21]: clustering or threshold-based methods, boundary detection methods, region growing methods, shape-based methods, and hybrid methods. In order to get the best result of cDNA microarray image we follow these steps:

(i) Image adjustment.

(ii) Segment spots from the background by thresholding (apply logarithmic transformation then threshold intensities)

(iii) Local thresholding

(iv) Combine the result of local threshold with global threshold.

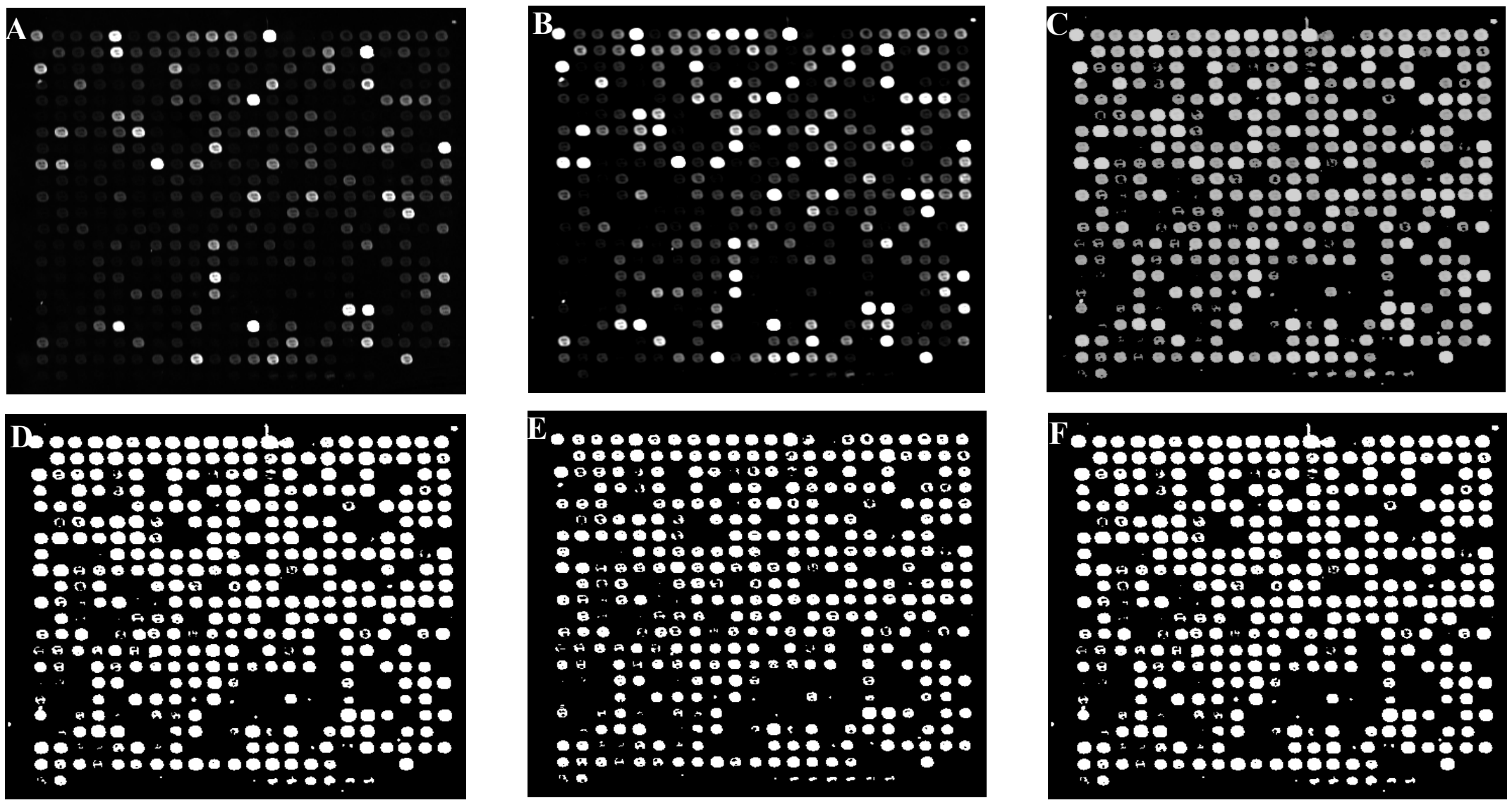

Fig. 5 The steps of segmentation A, Gabor filtered image, B, image adjustment, C, Logarithmic transformation, D, Global thresholding, E, Local thresholding, F, the combined image

## 3. DNA Matching by ANN

DNA sequence recognition involves knowing whether or not a new input has been presented before and stored in memory. The neural network classifier learning is an easy process because it is dependent only on the locally available information, but since the information is mixed together at the storage elements, unambiguous retrieval of stored information is not simple. The gradient descent is generalized to a non-linear, multi-layer feed forward network called back-propagation. This is a supervised learning algorithm that learns by first computing an error signal and then propagates the error backward through the network by assuming the network weights are the same in both the backward and forward directions. Back-propagation is a very popular and widely used network learning algorithm [22].

In this paper, we explain our work on training the cDNAs on the MDANN by using the back propagation algorithm and then matching the cDNA for recognition from the database. The ANN used here consists of 21 neurons (the input layer is of eight neurons, the hidden layer is of 12 neurons, and the output layer is of one neuron). Fig. 7 shows the structure of the ANN but the connections between the input and the hidden layer are not shown completely for the purpose of clarity and the complete structure is obtained by extending the connections to the entire image.

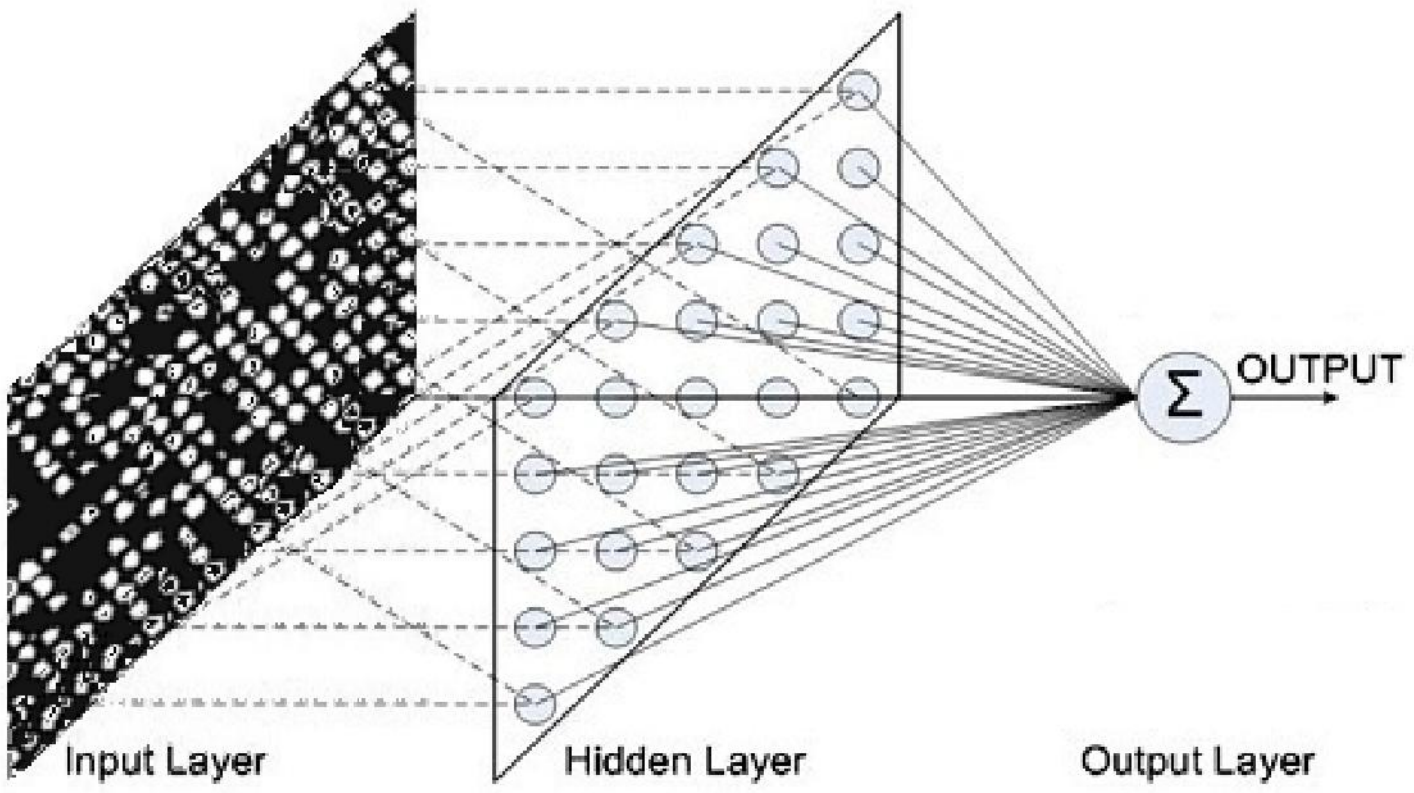

Fig.6 the structure of ANN

### 3.1 Algorithm for the ANN

Let $M_1$ and $M_2$ are structures containing $m$ layers and $k$ elements in each field and comprise of internal weights $W_{dl}, W_{dr}, W_{dl1}, W_{dr1}$, are matrices denoting the entire strengths of the image to neuron connection and the components $A_w, A_{dw}, A_b, A_d, A_{d1}$ are single row matrices denoting the weights of the connection of neurons with the image pixel points, their differential values, the bais values for the neurons, their differential values.

Step 1: calculation of the derivatives:

Now, let $i$ refers to the number cDNA we are trying to train on the network, $j$ refers to the number of regresses which is given by one in excess of the number of neurons, $k$ refers to the dimension of the input matrix, and $n$ refers to the number of neurons.

$$\delta = actual\ output - required\ output \tag{2}$$

For the weights eq. 3 and eq. 4

$$A_d(j) = A_d(j) + \delta, \qquad if \;\; j = 1 \tag{3}$$

$$A_d(j) = A_d(j) + \delta\left(\frac{2}{1+\exp(-2(M_{j1} x\ I(i) x M_{j2}} - 1\right), \qquad if \;\; j > 1 \tag{4}$$

Where the symbols $M_{j1}$ and $M_{j2}$ refer to elements in the $j^{th}$ layer elements of the structures $M_1$ and $M_2$ respectively and $I(i)$ refers to the $i^{th}$ matrix of cDNA input image to the network. The values are calculated for all values of $j$. Now $f$ refers to the activation function and is given by the eq. 5.

$$f = \left(\frac{-2}{1+\exp(-2(M_{j1} x\ I(i) x\ M_{j2})^2}\right) \tag{5}$$

For the internal weights eq. 5 and eq. 7.

$$W_{dl1}(j,k) = W_{dl1}(j,k) + (A_w(j) \times \delta \times \mathrm{f} \times (M'_{j2} \times I(i,k,:)')) \tag{6}$$

$$W_{dr1}(j,k) = W_{dr1}(j,k) + (A_w(j) \times \delta \times \mathrm{f} \times (M'_{j1} \times I(i,:,k))) \quad (7)$$

Where $W_{dl1}(j,k), W_{dr1}(j,k)$ refer to elements in the $j^{th}$ row and $k^{th}$ column of the matrices $W_{dl1}$ and $W_{dr1}$ respectively. $I(i,k,:)$ refers to the $k^{th}$ row elements of the $i^{th}$ input in the structure and $I(i,:,k)$ refers to the $k^{th}$ column elements of the $i^{th}$ input in the structure. These values are calculated for all the coordinates $(i,k)$.

For the bias values eq. 8.

$$A_{d1}(j) = A_{d1}(j) + (\delta \times f \times A_w(j)) \quad (8)$$

Step 2 : updating the weights and bias values and setting their derivatives back to zero:

For all values of j

For the weights eq. 9, eq. 10 and eq. 11.

$$A_w(j) = A_w(j) + (\alpha \times A_{dw}(j) - (\eta \times A_d(j)) \quad (9)$$

$$A_{dw}(j) = (\alpha \times A_{dw}(j) - (\eta \times A_d(j)) \quad (10)$$

$$A_d(j) = 0 \quad (11)$$

For the bias values eq. 12, eq. 13 and eq 14.

$$A_b(j) = A_b(j) + (\alpha \times A_{db}(j) - (\eta \times A_{d1}(j)) \quad (12)$$

$$A_{db}(j) = A_{db}(j) - (\eta \times A_{d1}(j)) \quad (13)$$

$$A_{dl}(j) = 0 \quad (14)$$

Where $\eta$ and $\alpha$ refer to the initial learning rate and the momentum factor respectively.

Step 3: updating the internal weights and setting the derivatives back to zero:

For all the coordinates

$$M_{jk1} = M_{jk1} + (\alpha \times W_{dl}(j,k)) - (\eta \times W_{dl1}(j,k)) \quad (15)$$

$$W_{dl}(j,k) = (\alpha \times W_{dl}(j,k)) - (\eta \times W_{dl1}(j,k)) \quad (16)$$

$$W_{dl1}(j,k) = 0 \quad (17)$$

$$M_{jk2} = M_{jk2} + (\alpha \times W_{dr}(j,k)) - (\eta \times W_{dr1}(j,k)) \quad (18)$$

$$W_{dr}(j,k) = (\alpha \times W_{dr}(j,k)) - (\eta \times W_{dr1}(j,k)) \quad (19)$$

$$W_{dr1}(j,k) = 0 \quad (20)$$

Where $M_{jk1}$ and $M_{jk2}$ refer to the $k^{th}$ element in the $j^{th}$ layer of the structure $M_1$ and $M_2$ respectively.

Step 4: calculating the output values after the training.

$$o = A_w(j) \; for \quad j = 1$$

$$o = A_w(1) + \sum_{j=2}^{n} A_w(j) \times \left( \frac{2}{1 + \exp((M_{j1} \times I(j) \times M_{j2}) + A_b(j))} \right) for \;\; j > 1 \quad (21)$$

The iteration goes on until the value calculated in Step 4 is the target value that was assigned to the image. This value is unique for every distinct microarray image and when a random microarray image is tested on the network, this value concludes as to whether the microarray image is already present in the database or not and if present the exact cDna to which it corresponds to. The ANN is trained by using 110 microarray image from [23] . During the training stage, each microarray image is assigned to a distinct arbitrary value. Now the trained ANN is subjected to a test to evaluate if the network is capable of recognizing a microarray image that have already trained on it and also to evaluate if it can reject a microarray image that have not trained on it. We will use some trained microarray images to test the ability of the ANN to identify the trained cDNA.

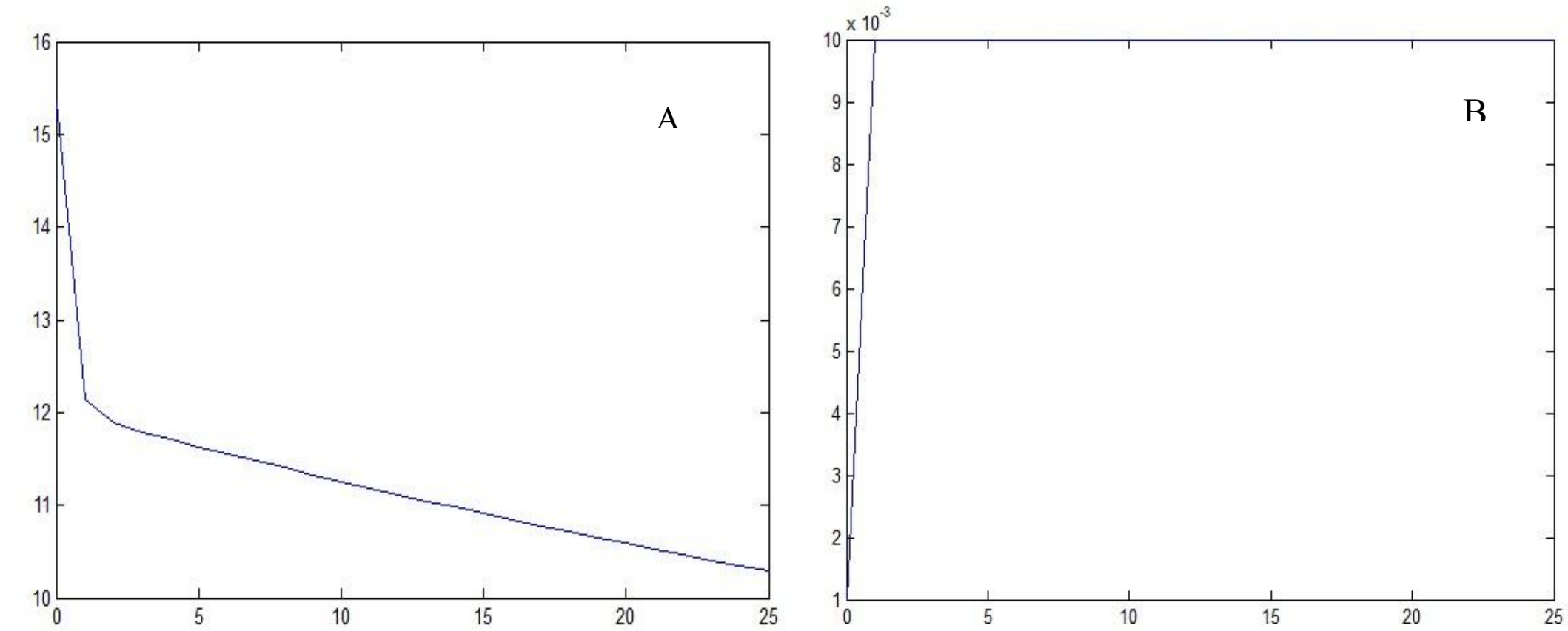

Fig.7 (A) the relation between the performance and the number of iterations,(B) the relation between the learning rate and the number of iterations

## 4. Result

The microarray image is enhanced in order to enable the proper extraction of gene spots. The segmented image which separates the background from foreground area is shown in Fig. 5 (a-c). Different enhancement operators are applied to microarray image like global and local thresholding as shown in Fig. 5(D-E). The combined microarray image Fig. 5 (F) is resized to 20x20. The resized image is trained on the ANN by using the gradient descent training procedure. The performance curve has been plotted against the number of iterations. It is observed that the performance is steadily improving and leading to a value around zero Fig. 6 (A). The rate of the learning curve has been shown Fig 6 (B), and this refers to the rate at which the network is learning. This rate is altered continuously in the algorithm every time the network is subject to iteration.

## 5. Conclusion

In this paper, we present a new method for cDNA recognition based on the artificial neural network (ANN). Microarray imaging is used for the concurrent identification of thousands of genes. We have segmented the location of the spots in a cDNA microarray images. The segmented cDNA microarray image is resized and it is used as an input for the proposed artificial neural network. For matching and recognition, we have trained the artificial neural network. Recognition results are given for the galleries of cDNA sequences. The numerical results show that, the proposed matching technique is an effective in the cDNA sequences process. We also compare our results with previous results and find out that, the proposed technique is an effective matching performance.

## Acknowledgement

The authors would like to thank the associate editor and the anonymous reviewers for their constructive comments.